\documentclass[10pt,twocolumn,letterpaper]{article}

\usepackage{cvpr}
\usepackage{times}
\usepackage{epsfig}
\usepackage{graphicx}
\usepackage{amsmath}
\usepackage{amssymb}
\usepackage{slashbox,multirow}

\usepackage[pagebackref=true,breaklinks=true,letterpaper=true,colorlinks,bookmarks=false]{hyperref}

\newcommand*{\affaddr}[1]{#1} 
\newcommand*{\affmark}[1][*]{\textsuperscript{#1}}
\newcommand*{\email}[1]{\texttt{#1}}

\cvprfinalcopy 

\def\cvprPaperID{5632} 

\ifcvprfinal\fi
\begin{document}

\title{StereoGAN: Bridging Synthetic-to-Real Domain Gap \\by Joint Optimization of Domain Translation and Stereo Matching}

\author{%
Rui Liu\affmark[1] \quad Chengxi Yang\affmark[2] \quad Wenxiu Sun\affmark[2] \quad Xiaogang Wang\affmark[1] \quad Hongsheng Li\affmark[1]\\
\affaddr{\affmark[1]CUHK-SenseTime Joint Laboratory, Chinese University of Hong Kong \qquad \affmark[2]SenseTime Research}\\
\email{ruiliu@link.cuhk.edu.hk} \quad \email{\{yangchengxi, sunwenxiu\}@sensetime.com}\\
\email{\{xgwang, hsli\}@ee.cuhk.edu.hk}\\
}

\maketitle

\begin{abstract}
   Large-scale synthetic datasets are beneficial to stereo matching but usually introduce known domain bias. Although unsupervised image-to-image translation networks represented by CycleGAN show great potential in dealing with domain gap, it is non-trivial to generalize this method to stereo matching due to the problem of pixel distortion and stereo mismatch after translation.
   In this paper, we propose an end-to-end training framework with domain translation and stereo matching networks to tackle this challenge.
   First, joint optimization between domain translation and stereo matching networks in our end-to-end framework makes the former facilitate the latter one to the maximum extent.
   Second, this framework introduces two novel losses, i.e., bidirectional multi-scale feature re-projection loss and correlation consistency loss,
   to help translate all synthetic stereo images into realistic ones as well as maintain epipolar constraints.
   The effective combination of above two contributions leads to impressive stereo-consistent translation and disparity estimation accuracy.
   In addition, a mode seeking regularization term is added to endow the synthetic-to-real translation results with higher fine-grained diversity.
   Extensive experiments demonstrate the effectiveness of the proposed framework on bridging the synthetic-to-real domain gap on stereo matching.
\end{abstract}

\section{Introduction}
\label{introduction}
With the fast development of deep neural networks~\cite{alexnet, resnet} and large-scale benchmarks~\cite{Ros_2016_CVPR, HernandezBMVC17, kittidataset}, deep learning-based stereo matching methods have made great progress in the past decade~\cite{dispnet, gcnet}. These methods, however, relying on a large quantity of high-quality \emph{left-right-disparity} training data. Although the input images to the stereo matching networks ( \emph{i.e.}, left and right images) are relatively easy to collect using stereo rigs in the real world, their corresponding ground-truth disparities are very difficult to collect. Instead, researchers tend to create synthetic training datasets~\cite{dispnet, Ros_2016_CVPR, HernandezBMVC17} with perfect disparities. In this way, the demand of large quantity of training data is alleviated. However, the non-negligible domain gaps between synthetic and real must be considered when generalizing to real domains.
In order to mitigate the domain gaps, some of the previous works~\cite{chang2018pyramid,Tonioni_2019_realadapt} train their models in two stages. Firstly the model is trained on synthetic dataset and then fine-tuned on a particular real dataset in either supervised~\cite{Pang2017iccvw, chang2018pyramid, guo2019group} or unsupervised manner~\cite{Tonioni_2017_ICCV, Tonioni_2019_learn2adapt}. In this paper, we focus on the latter one, a more challenging task with no ground-truth for the real target-domain data.
\begin{figure}[t]
    \centering
    \includegraphics[width=0.99\linewidth]{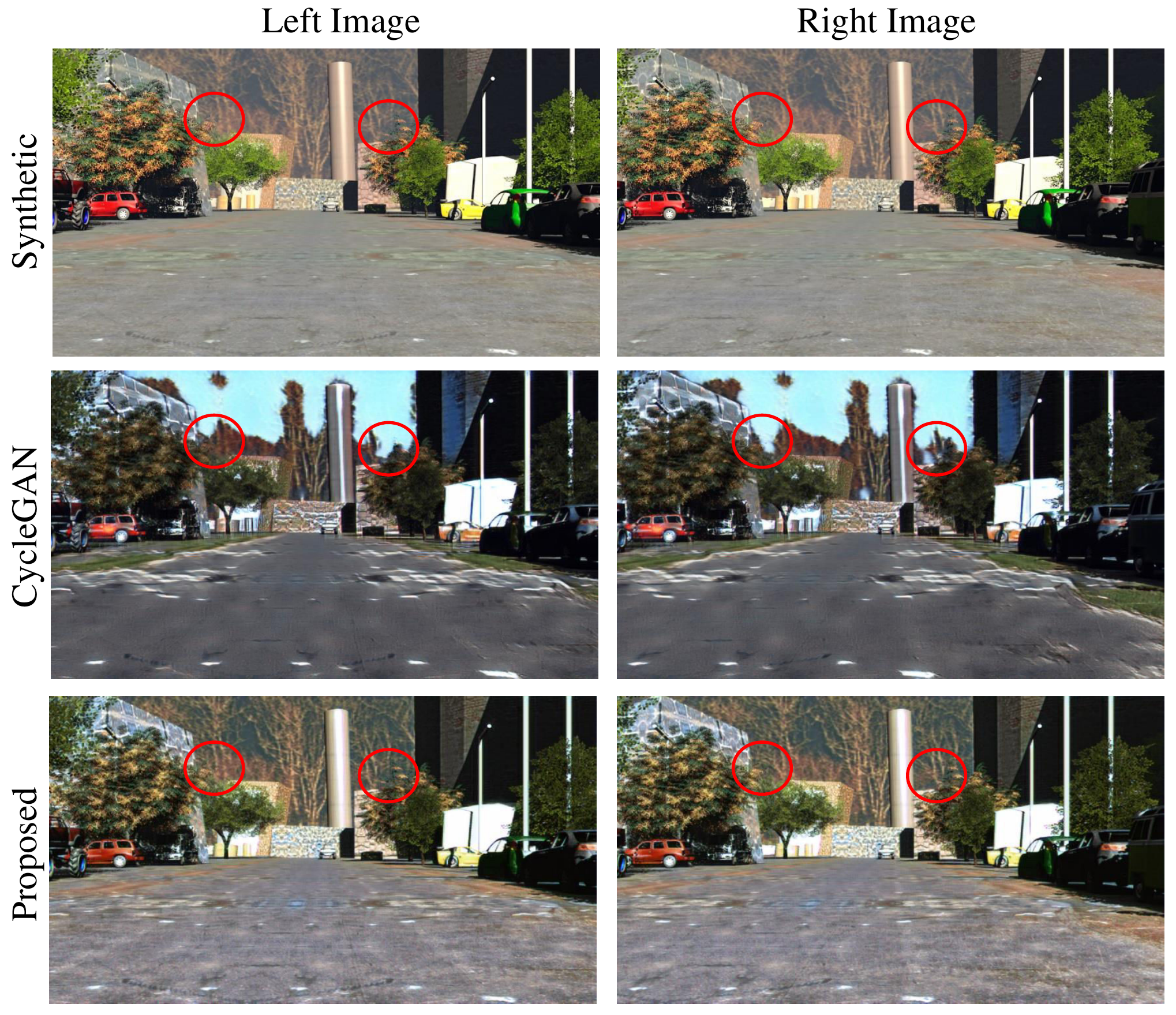}
    \caption{Domain translation results. Top row: stereo images from synthetic domain. Middle row: synthetic-to-real translated results by CycleGAN. Bottom row: synthetic-to-real translated results by our proposed model.}
    \label{fig1}
\end{figure}

Existing unsupervised online adaptation methods advanced the research progress, however, still have difficulties on handling the domain gaps between source and target domains~\cite{Tonioni_2017_ICCV, Tonioni_2019_learn2adapt}.
Moreover, these methods introduce extra computation compared to a feed-forward neural network, although they have striven to reduce the computation complexity of updating network parameters~\cite{Tonioni_2019_realadapt}.

Recently, unsupervised image-to-image translation models achieved great success~\cite{CycleGAN2017,NIPS2017_6672,DRIT} and thus were adopted in domain adaptation methods to tackle many applications such as semantic segmentation, person re-identification and object detection~\cite{cycada,Tsai_2018_CVPR, Sankaranarayanan_2018_CVPR, Deng_2018_CVPR}. However, it is non-trivial to generalize this series of methods to stereo matching. The middle row of Figure~\ref{fig1} reveals two main challenges for translation in stereo matching. 1) The general image-to-image translation does not take epipolar constraints into consideration, which leads to inconsistent textures and thus ambiguity of disparity, as emphasized by red circles.
2) It only attempts to transfer domain styles while neglecting the fact that its purpose should be serving the stereo matching networks. For instance, since most background of our synthetic images is brown mountains while that of real images in the training set is blue sky, the vanilla CycleGAN~\cite{CycleGAN2017} regards this to be domain style and tries to translate from brown mountains to blue sky as shown in the first two rows of Figure~\ref{fig1}.
This would confuse stereo matching network, because the useful textures for stereo matching in the sky is definitely much less than those in the mountains.
In this paper, we successfully addressed these two challenges by properly designed stereo constraints and joint training scheme. The intermediate image translation results are shown in the bottom row of Figure~\ref{fig1}.

In particular, we propose an end-to-end deep learning framework consisting of domain translation and stereo matching networks to estimate stereo disparity on the target domain, using only source-domain synthetic stereo image pairs with ground-truth disparity and target-domain real stereo image pairs without any annotation.
The stereo image translation is constrained by a novel bidirectional multi-scale feature re-projection loss and a correlation consistency loss.
The former one is realized by a multi-scale feature re-projection module. For feature maps at each layer of domain translation networks, the inverse warping~\cite{STN} of the right feature map according to the given disparity should be as close as its corresponding left feature map. Both ground-truth disparity for synthetic data and estimated disparity for real data would contribute to joint training in a bidirectional manner.
We also introduce a correlation consistency loss to ensure that the reconstructed stereo images should maintain consistent correlation feature maps, which are extracted from the stereo matching network, with those original images.

\begin{figure}[t]
    \centering
    \includegraphics[width=0.99\linewidth]{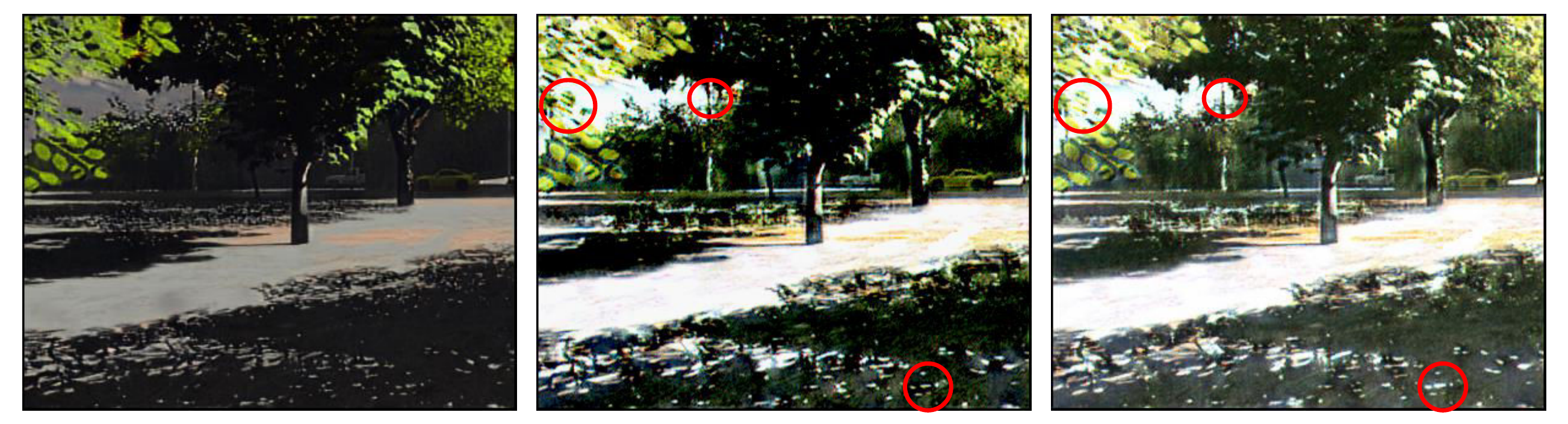}
    \caption{The effect of mode seeking regularization term. Leftmost image is from synthetic domain, and middle image and rightmost image are translated from leftmost image with different random maps. Red circles emphasize the fine-grained difference between middle image and rightmost image. Please zoom in to observe more details. }
\label{modeseekfig}
\end{figure}

In addition, we observed that real stereo pairs usually do not exactly match each other due to different camera configurations and settings. To this end, inspired by successful applications of using noise to manipulate image~\cite{stylegan, Mao_2019_CVPR}, we propose a mode seeking regularization term to ensure the fine-grained diversity in synthetic-to-real translation,
as shown in Figure~\ref{modeseekfig}. As we could observe as circled in red, the local intensity between the left image and right image varies, which simulates the real data. With such augmentation, the domain translation makes the stereo matching in the real domain more robust and effective.


In summary, our contributions are listed as follows:
\begin{itemize}
  \item We for the first time combine unsupervised domain translation with disparity estimation in an end-to-end framework to tackle the challenging problem of stereo matching in the absence of real ground-truth disparities.

  \item We propose novel stereo constraints including the bidirectional multi-scale feature re-projection loss and the correlation consistency loss, which better regularizes this joint framework to achieve stereo-consistent translation and accurate stereo matching. The additional mode seeking regularization endows the synthetic-to-real translation with higher fine-grained diversity.

  \item Extensive experiments demonstrate that our proposed model outperforms the state-of-the-art unsupervised adaptation approaches for stereo matching.
\end{itemize}

\section{Related Work}
\textbf{Stereo matching} conventionally follows a four-step pipeline including matching cost computation, cost aggregation, disparity optimization and post-processing~\cite{taxostereo}. Local descriptors such as absolute difference (AD), sum of squared difference (SAD) and so on are usually adopted for measuring left-right inconsistency, so as to calculate matching costs for all possible disparities. Cost aggregation and disparity optimization are usually treated as a $2$D graph partitioning problem, which could be optimized by graph cut~\cite{graphcut} or belief propagation~\cite{SunBelief,KlausBelief}. Semi-global matching (SGM)~\cite{SGM} approximates the global optimization with dynamic programming.

Deep learning-based stereo matching methods have achieved great progress due to the rise of deep neural networks~\cite{alexnet, resnet} and large-scale benchmarks~\cite{geiger2012we, kittidataset} in the last decade. Among them, Zbontar and LeCun~\cite{comp_stereo} for the first time presented the computation of stereo matching costs by a deep Siamese network. Luo \etal~\cite{Luo2016EfficientDL} accelerated the computation of matching costs by correlating unary features.
Recently, many end-to-end neural networks were developed to directly predict the whole disparity maps from stereo image pairs~\cite{dispnet, Pang2017iccvw, EdgeStereo, yang2018segstereo, gcnet, chang2018pyramid, Yu2018DeepSM, guo2019group}. Among them, DispNet~\cite{dispnet} is a pioneer work which for the first time uses an end-to-end deep learning framework to directly regress disparity maps. The follow-up work GCNet~\cite{gcnet} introduces $3$D convolutional networks to aggregate contextual information for obtaining better cost volumes.

\textbf{Domain adaptation}
methods have shown great potential in filling the gap between synthetic and real domains. Previous works attempted to solve this problem by either learning domain-invariant representations~\cite{pmlr-v37-ganin15, JMLR:v17:15-239} or pushing two domain distributions to be close~\cite{Gretton2012,Tzeng2014DeepDC,coral,deepcoral}. For example, the gap between source and target domain could be filled by matching the distribution~\cite{MMD,LongTransfer} or statistics~\cite{coral,deepcoral} of deep features.

Recently, unsupervised image-to-image translation models achieved great success under unpaired setting~\cite{CycleGAN2017,NIPS2017_6672,DRIT} and thus were applied as domain adaptation methods in many applications including semantic segmentation, person re-identification and object detection~\cite{cycada,Tsai_2018_CVPR, Sankaranarayanan_2018_CVPR, Deng_2018_CVPR}.

In the field of stereo matching, unsupervised online adaptation advanced great progress. These methods first train a disparity estimation network on synthetic data and then fine-tune it online using unsupervised loss such as re-projection loss when continuously accessing new stereo pairs from other domains~\cite{Tonioni_2017_ICCV, Tonioni_2019_realadapt}. This unsupervised adaptation strategy is then incorporated in a meta-learning framework~\cite{Tonioni_2019_learn2adapt}.

\section{Method}
Given a set of $N$ synthetic \emph{left}-\emph{right}-\emph{disparity} tuples $\{(x_l, x_r, x_d)_i\}_{i=1}^{N}$ in the source domain $\cal X$, where $(x_l, x_r, x_d) \in (\mathcal{X}_L, \mathcal{X}_R, \mathcal{X}_D) = \mathcal{X}$, and a set of $M$ real stereo images $\{(y_l, y_r)\}_{j=1}^{M}$ in the target domain $\cal Y$ without any ground-truth disparity, where $(y_l, y_r) \in (\mathcal{Y}_L, \mathcal{Y}_R)$, our goal is to learn an accurate disparity estimation network $F$ for estimating the disparity $\hat{y}_d = F(y_l, y_r)$ on the target domain.

For the sake of clear formulation, we define a paired set $(\mathcal{X}_L, \mathcal{X}_R)=\{(x_{l1}, x_{r1}), (x_{l2}, x_{r2}), ..., (x_{lN}, x_{rN})\}$ where $(x_{li}, x_{ri})$ stands for a paired stereo image, \textit{i.e.}, a left image $x_{li}$ and its corresponding right image $x_{ri}$ (see Eqs. (\ref{eq:smloss}-\ref{eq:corrloss})). We also define an unpaired set $\{\mathcal{X}_L, \mathcal{X}_R\}=\{x_{l1}, x_{r1}, x_{l2}, x_{r2}, ..., x_{lN}, x_{rN}\}$ where we can only sample a single left or right image (see Eqs. (\ref{eq:advloss}-\ref{eq:cycloss})).

\begin{figure*}[t]
    \centering
    \includegraphics[width=0.99\linewidth]{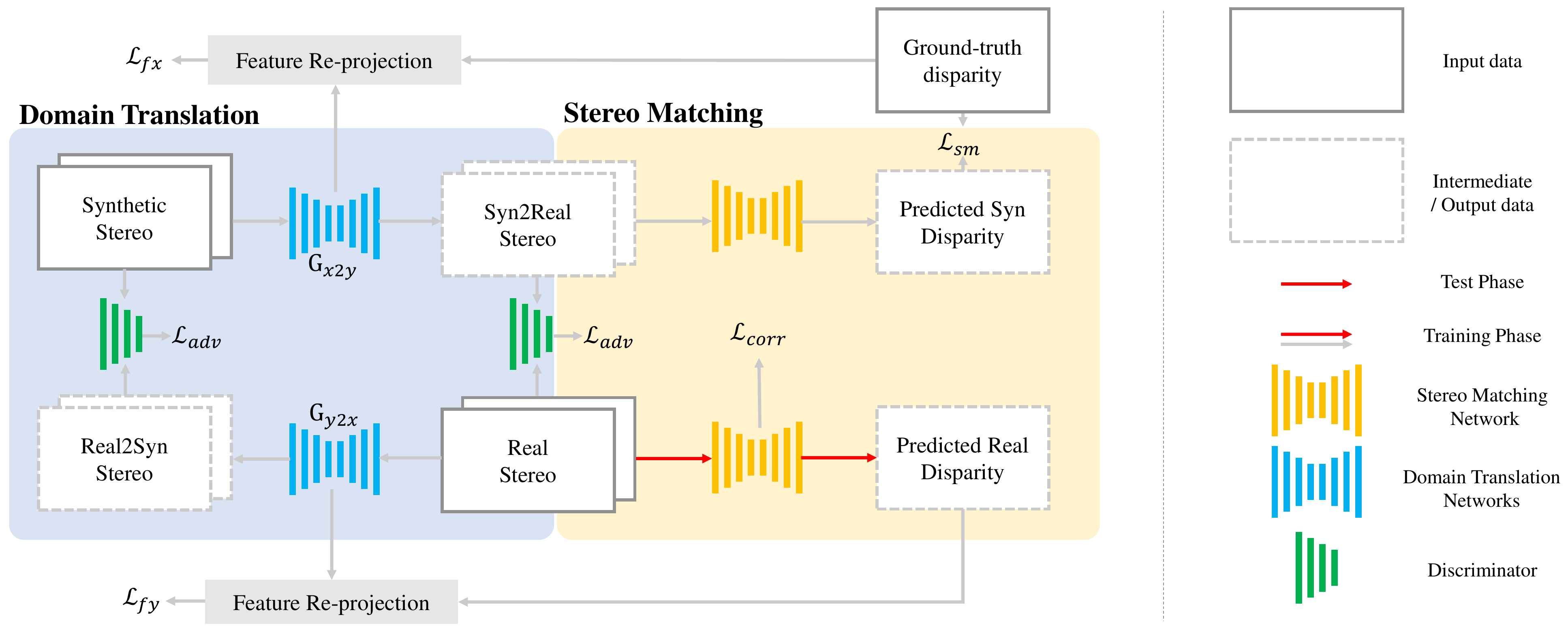}
    \caption{The joint framework of our proposed method. Blue-background block shows our domain translation component and orange-background block shows our stereo matching component. Different blocks, lines and nets are labeled in the rightmost of this figure. $F$ denotes the stereo matching network. Note that we omit cycle consistency loss due to the limited space. }
    \label{fig:framework}
\end{figure*}

Different from previous works that directly train stereo matching network $F$ with synthetic data~\cite{dispnet, gcnet, Tonioni_2019_realadapt}, we propose a joint domain translation and stereo matching framework, which aims to translate synthetic-style stereo images into realistic ones with novel stereo constraints and thus better cooperate with the stereo matching network in an end-to-end manner, as shown in Figure~\ref{fig:framework}.

\subsection{Cycle-consistency Domain Translation for Stereo Matching}
\noindent \textbf{Cycle-consistency domain translation loss}. To help synthetic-to-real translation network $G_{x2y}$ capture the global domain style of the real datasets, we adopt a real domain discriminator $D_y$ whose goal is to distinguish synthetic-to-real generated images from real-domain images. On the contrary, $G_{x2y}$ learns to generate images that look similar to real-domain images to fool the real domain discriminator $D_y$. These two sub-nets constitute a minimax game that optimizes in an adversarial manner and achieves optimal when $D_y$ cannot tell whether images are generated or not. The adversarial loss for synthetic-to-real generation is formulated as:
\begin{equation}
\label{eq:advloss}
\begin{aligned}
  \mathcal{L}_{adv} & (G_{x2y}, D_{y}, \mathcal{X}, \mathcal{Y}) = \mathbb{E}_{y \sim \{\mathcal{Y}_L, \mathcal{Y}_R\}} \left[ \log{D_y(y)} \right] \\
  & + \mathbb{E}_{x \sim \{\mathcal{X}_L, \mathcal{X}_R\}} \left[ \log{ (1 - D_y(G_{x2y}(x)) } \right],
\end{aligned}
\end{equation}
where ${y \sim \{\mathcal{Y}_L, \mathcal{Y}_R\}}$ means a single real image $y$ is sampled from the non-paired real-domain set $\{\mathcal{Y}_L, \mathcal{Y}_R\}$. We also introduce a similar adversarial loss for supervising the process of real-to-synthetic generation as $\mathcal{L}_{adv} (G_{y2x}, D_{x}, \mathcal{Y}, \mathcal{X})$.

Adversarial losses could only supervise $G_{x2y}$ and $G_{y2x}$ to produce images that are not distinguishable by domain discriminators, but any random permutation of outputs can happen without any other constraints. In order to regularize $G_{x2y}$ and $G_{y2x}$ to be one-to-one mapping, the cycle consistency loss is also adopted,
\begin{equation}
\label{eq:cycloss}
\begin{aligned}
  & \quad \mathcal{L}_{cyc} (G_{x2y}, G_{y2x})  \\
  & = \mathbb{E}_{y \sim \{\mathcal{Y}_L, \mathcal{Y}_R\}} \left[ \left\|G_{x2y}(G_{y2x}(y)) - y \right\|_1 \right] \\
  & + \mathbb{E}_{x \sim \{\mathcal{X}_L, \mathcal{X}_R\}} \left[ \left\|G_{y2x}(G_{x2y}(x)) - x \right\|_1 \right].
\end{aligned}
\end{equation}

To sum up, the cycle-consistency domain translation loss following the CycleGAN~\cite{CycleGAN2017} can be defined as
\begin{equation}
\label{eq:CDTloss}
\begin{aligned}
  &\mathcal{L}_{cdt} (G_{x2y}, G_{y2x}, D_x, D_y) =  \mathcal{L}_{adv}(G_{x2y}, D_{y}, \mathcal{X}, \mathcal{Y})  \\
  &+ \mathcal{L}_{adv}(G_{y2x}, D_{x}, \mathcal{Y}, \mathcal{X}) + \lambda_{cyc} \mathcal{L}_{cyc} (G_{x2y}, G_{y2x}).
\end{aligned}
\end{equation}

\noindent \textbf{Stereo matching loss}. Since our goal is to learn a mapping from real-domain stereo image to disparity map with only annotated synthetic stereo images and unlabeled real ones, it is straight-forward to take advantage of the results of synthetic-to-real translation. Given a paired synthetic tuple $(x_l, x_r, x_d)$, we argue that the translated stereo pair $(G_{x2y}(x_l), G_{x2y}(x_r))$ could be regarded as real-domain images and such translated stereo pair should match its ground-truth disparity $x_d$. Therefore, we formulate the stereo matching loss as:
\begin{equation}
\label{eq:smloss}
\begin{aligned}
  & \mathcal{L}_{sm} (F) = \mathbb{E}_{(x_l,x_r,x_d) \sim \mathcal{X}} \left[ \left\|F(G_{x2y}(x_l),G_{x2y}(x_r)) - x_d \right\|_1 \right],
\end{aligned}
\end{equation}
where $F(\cdot, \cdot)$ is the stereo matching network for estimating disparities from real-domain stereo images.

These two losses construct a simple framework that optimizes stereo matching network with the assistance of domain translation networks. However, it may introduce the problem of pixel distortion and stereo mismatch during translation.

\subsection{Joint Domain Translation and Stereo Matching}
To tackle the above mentioned challenges, we should ensure that domain translation networks only transfer global domain style while maintain the epipolar consistency, which contributes to the improvement of stereo matching. To achieve this, we propose a joint optimization scheme between domain translation and stereo matching with novel constraints.

\begin{figure}[t]
    \centering
    \includegraphics[width=0.98\linewidth]{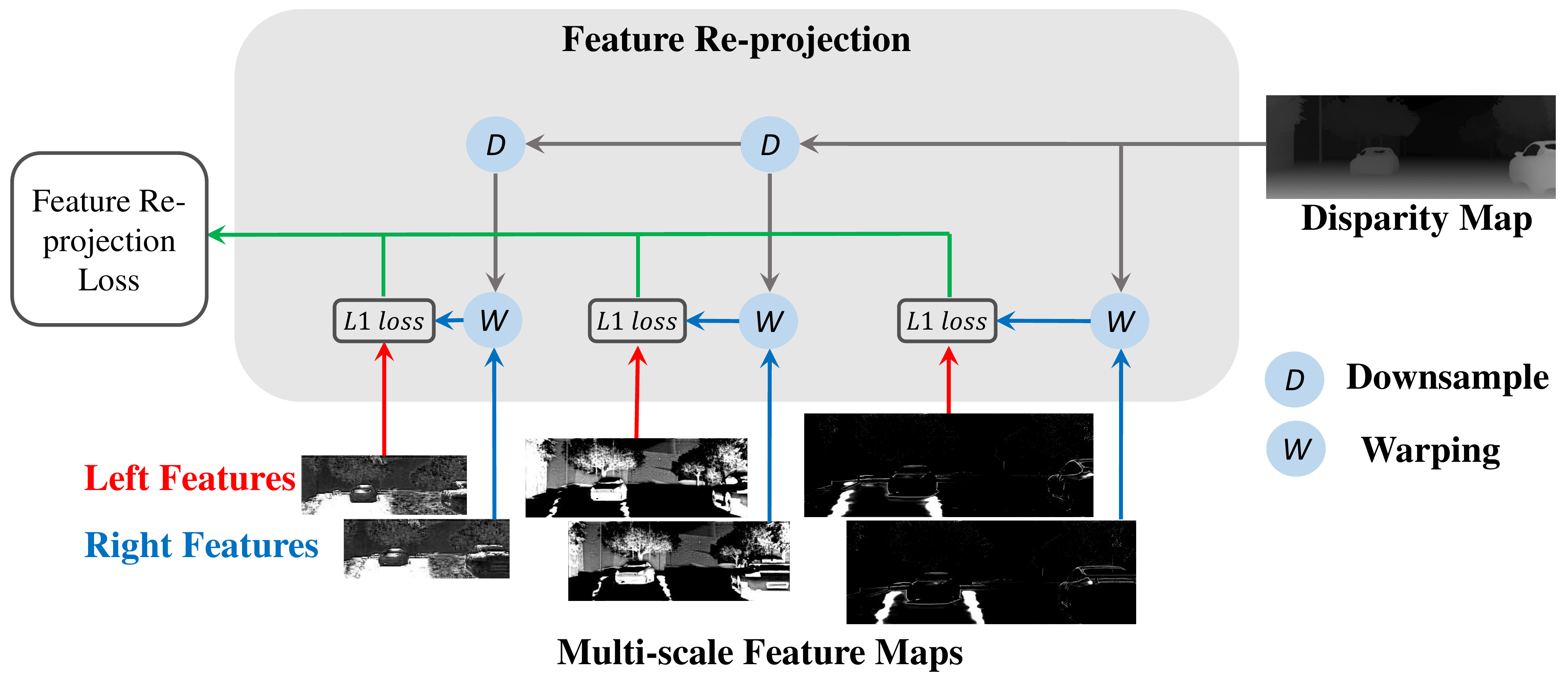}
    \caption{Detailed structure of feature re-projection module. This figure demonstrates the calculating process of feature re-projection loss for synthetic data with ground-truth disparity. Note that stereo matching networks usually output multi-scale disparities, so we remove downsample function when dealing with real data.}
    \label{fig:warping}
\end{figure}

Before diving into novel constraints, we would first introduce our newly-proposed multi-scale feature re-projection module, which establishes a bidirectional connection between domain translation component and stereo matching component by left-right consistency check, as illustrated in Figure~\ref{fig:warping}. For each intermediate layer of domain translation networks, the inversely warped right feature map should be the same as its corresponding left feature map. This inverse warping operation is completed with properly downsampled disparity map using differentiable bilinear sampling technique~\cite{STN}. Note that the given disparity could be either ground-truth one for synthetic stereo or estimated one for real stereo, which calculate feature re-projection loss for synthetic or real stereo images respectively. The former endows the domain translation networks with strong epipolar constraints while the latter provides extra supervision for training stereo matching network.

\noindent \textbf{Feature re-projection loss for synthetic images}. We argue that the intermediate feature maps for generating the domain-translated left and right images should be the same at 3D physical locations.
To model this constraint, we utilize synthetic ground-truth disparity to warp the intermediate feature maps of both $G_{x2y}$ and $G_{y2x}$ along the synthetic-real-synthetic cycle translation. If the stereo image pairs are well translated, the inversely warped right feature map should match the left feature exactly.
The feature re-projection loss for synthetic images is formulated as
\begin{equation}
\label{eq:fwxloss}
\begin{aligned}
  & \quad \mathcal{L}_{fx} (G_{x2y}, G_{y2x})  \\
  & = \mathbb{E}_{(x_l,x_r,x_d) \sim \mathcal{X}} \frac{1}{T_1} \sum_{i=1}^{T_1} \left[ \left\|W(G_{x2y}^{(i)}(x_r), x_d) - G_{x2y}^{(i)}(x_l) \right\|_1 \right . \\
  & + \left . \left\| W(G_{y2x}^{(i)}(G_{x2y}(x_r)), x_d) - G_{y2x}^{(i)}(G_{x2y}(x_l)) \right\|_1 \right],
\end{aligned}
\end{equation}
where $T_1$ is the total number of layers of translation networks, $G^{(i)}(x)$ denotes the feature of image $x$ at $i$th-layer the translation network $G$, the inverse warping function $W(G^{(i)}(x_r), x_d)$ warps the right feature map $G^{(i)}(x_r)$ with the ground-truth disparity $x_d$.

\noindent \textbf{Feature re-projection loss for real images}. For a general stereo matching network such as DispNet~\cite{dispnet}, it naturally outputs multi-scale disparities, which can be formed from correlation features at different neural network layers.
These multi-scale disparity maps can be used to warp the intermediate feature maps for both $G_{x2y}$ and $G_{y2x}$ along the real-synthetic-real cycle translation. Then the $L1$ distance between the left feature and the inversely warped right feature provides an extra supervision for updating the parameters of disparity estimation network $F$. This loss could be formulated as
\begin{equation}
\label{eq:fwyloss}
\begin{aligned}
  & \quad \mathcal{L}_{fy} (F)  \\
  & = \mathbb{E}_{(y_l,y_r) \sim (\mathcal{Y}_L, \mathcal{Y}_R)} \frac{1}{T_1} \sum_{i=1}^{T_1} \left[ \left\|W(G_{y2x}^{(i)}(y_r), \hat{y}_d) - G_{y2x}^{(i)}(y_l) \right\|_1 \right . \\
  & + \left . \left\| W(G_{x2y}^{(i)}(G_{y2x}(y_r)), \hat{y}_d) - G_{x2y}^{(i)}(G_{y2x}(y_l)) \right\|_1 \right],
\end{aligned}
\end{equation}
where 
$\hat{y}_d$ is the estimated disparity of real stereo image pairs by $F(y_l, y_r)$.

Different from previous works which directly warp images at the origin scale~\cite{garg2016unsupervised, zhao2019geometry}, our warping operation is based on multi-scale feature maps. Since features at different layers model image structures of different scales, this constraint could help supervise the training of stereo matching network from multiple scales (from global to local regions), leading to impressive improvement on disparity estimation accuracy. In addition, it leaves some space for fine-grained noise modeling upon pixel level (see Figure~\ref{modeseekfig}), which would be introduced in the mode seeking regularization term, described later in this section.

\noindent \textbf{Correlation consistency loss}. Feature re-projection losses may not totally address the stereo-mismatch issue yet. Since there is no ground-truth disparity for real-domain stereo images, warping features with estimated disparity may introduce some bias into the joint framework. For example, the value of $\mathcal{L}_{fy}$ for a certain \emph{left}-\emph{right}-\emph{disparity} tuple may be $0$, but it still makes a limited effect on stereo matching, even makes a negative effect. This is because the phenomenon of pixel distortion during domain translation and inaccurate estimation during stereo matching occur simultaneously.

To reduce such impact, stereo matching network is utilized to supervise both $G_{y2x}$ and $G_{x2y}$ along the real-synthetic-real cycle translation. We denote the reconstructed real image by such cycle translation as $y'=G_{x2y}(G_{y2x}(y))$ for ease of presentation.
Given a pair of real stereo images $(y_l,y_r)$, we could obtain their reconstructed pair $(y_l',y_r')$. The correlation features of $(y_l',y_r')$ from each layer of stereo matching network should match those of $(y_l,y_r)$. In addition, we make a cross-pair for constructing a tighter loss, which is calculated by pushing correlation features of both $(y_l',y_r)$ and $(y_l,y_r')$ to be close to those of $(y_l,y_r)$.
Therefore, we formulate this constraint for real-domain images as the correlation consistency loss between multi-layer correlation features:
\begin{equation}
\label{eq:corrloss}
\begin{aligned}
  & \quad \mathcal{L}_{corr} (G_{x2y}, G_{y2x})  \\
  & = \mathbb{E}_{(y_l,y_r) \sim (\mathcal{Y}_L, \mathcal{Y}_R)} \frac{1}{T_2} \sum_{i=1}^{T_2} \left[ \left\|F^{(i)}(y_l', y_r) - F^{(i)}(y_l, y_r) \right\|_1 \right. \\
  & + \left. \left\|F^{(i)}(y_l, y_r') - F^{(i)}(y_l, y_r) \right\|_1 \right. \\
  & + \left. \left\|F^{(i)}(y_l', y_r') - F^{(i)}(y_l, y_r) \right\|_1 \right],
\end{aligned}
\end{equation}
where $T_2$ is the total number of correlation aggregation layers which are after the individual image feature encoding layers and $F^{(i)}(y_l,y_r)$ denotes the correlation aggregation feature of the stereo pair $(y_l,y_r)$ at $i$th-layer of the stereo matching network $F$.



\begin{table*}
\begin{center}
\begin{tabular}{c|c|c|c|c|c|c|c|c|c|c|c|c}
\hline
\multirow{2}{*}{Dataset} & \multirow{2}{*}{Method} & \multicolumn{2}{c|}{D1-all (\%)} & \multicolumn{2}{c|}{EPE} & \multicolumn{2}{c|}{\textgreater 2px (\%)} & \multicolumn{2}{c|}{\textgreater 4px (\%)}  & \multicolumn{2}{c|}{\textgreater 5px (\%)}  & Time  \\
\cline{3-12}
& & Noc & All & Noc & All & Noc & All & Noc & All & Noc & All & (s) \\
\hline
& Inference & $10.75$ & $11.14$ & $1.817$ & $1.961$ & $20.52$ & $20.86$ & $8.40$ & $8.85$ & $5.68$ & $6.06$ & $0.06$ \\
Synthia& SL+Ad~\cite{Tonioni_2019_realadapt} & $10.02$ & $10.58$ & $1.596$ & $1.724$ & $19.86$ & $20.16$ & $7.98$ & $8.42$ & $5.53$ & $5.82$ & $0.19$ \\
to& L2A+Wad~\cite{Tonioni_2019_learn2adapt} & $9.88$ & $10.48$ & $1.569$ & $1.697$ & $17.32$ & $17.70$ & $6.78$ & $7.12$ & $5.01$ & $5.54$ & $0.23$ \\
KITTI2015& CycleGAN & $10.20$ & $10.69$ & $1.653$ & $1.890$  & $17.83$ & $18.15$ & $6.83$ & $7.39$ & $5.10$ & $5.65$ & $0.06$ \\
\cline{2-13}
& Proposed & $\textbf{8.78}$ & $\textbf{9.26}$ & $\textbf{1.488}$ & $\textbf{1.631}$ & $\textbf{15.74}$ & $\textbf{16.09}$ & $\textbf{5.73}$ & $\textbf{6.17}$ & $\textbf{4.55}$ & $\textbf{5.08}$ & $0.06$ \\
\hline
\hline
& Inference & $52.65$ & $53.07$ & $9.351$ & $9.513$ & $63.95$ & $64.30$ & $45.07$ & $45.52$ & $39.33$ & $39.79$ & $0.06$ \\
Driving& SL+Ad~\cite{Tonioni_2019_realadapt} & $39.16$ & $39.49$ & $4.698$ & $4.775$ & $53.33$ & $53.61$ & $30.22$ & $30.56$ & $24.18$ & $24.52$ & $0.19$ \\
to& L2A+Wad~\cite{Tonioni_2019_learn2adapt} & $26.33$ & $26.90$ & $2.878$ & $3.017$ & $40.59$ & $41.57$ & $17.31$  & $18.01$ & $12.55$ & $13.27$ & $0.23$ \\
KITTI2015& CycleGAN  & $31.23$ & $31.74$ & $3.272$ & $3.444$ & $44.34$ & $45.29$ &  $19.76$ & $20.34$ & $15.08$ & $15.68$ & $0.06$ \\
\cline{2-13}
& Proposed & $\textbf{25.18}$ & $\textbf{25.71}$ & $\textbf{2.584}$ & $\textbf{2.752}$ & $\textbf{39.16}$ & $\textbf{40.24}$ & $\textbf{15.83}$ & $\textbf{16.55}$ & $\textbf{11.04}$ & $\textbf{11.60}$ & $0.06$ \\
\hline
\end{tabular}
\end{center}
\caption{Evaluation results of the proposed method compared to different methods on Synthia-to-KITTI2015 and Driving-to-KITTI2015. Lower value means better performance. }
\label{table:comparison}
\end{table*}

\noindent \textbf{Mode seeking loss}. The above losses could well maintain the stereo consistency of the domain-translated images.
However, in practice, the stereo images also show slight variations between the left and right images, because of sensor noise, different camera configurations, etc. To model such left-right image variations,
we propose a mode seeking regularization term following \cite{Mao_2019_CVPR} to make the generators create small but realistic variations between the generated left and right images, as demonstrated in Figure~\ref{modeseekfig}. A Gaussian random map $z$ is introduced into the synthetic-to-real translation networks $G_{x2y}(x,z)$ to model the variations of the generated images. When training domain translation networks, we attempt to maximize the $L1$ distance between two generated outputs from the same original image $x$ with two different random maps $z_1$ and $z_2 \sim p(z)$, where $p(z)$ denotes a prior Gaussian distribution with zero mean and unity variance.
Since this term has no optimal point, we linearly decay its weight to zero during training.
This loss is formulated as
\begin{equation}
\label{eq:msloss}
\begin{aligned}
  & \quad \mathcal{L}_{ms} (G_{x2y})  \\
  & = \mathbb{E}_{x \sim \{\mathcal{X}_L, \mathcal{X}_R\}, z_1,z_2 \sim p(z)} \left[ \frac{\left\| z_1-z_2 \right\|_1}{ \left\|G_{x2y}(x,z_1)-G_{x2y}(x,z_2)\right\|_1 } \right].
\end{aligned}
\end{equation}

\subsection{Full Objective and Optimization}
Putting all the losses introduced above into an overall objective function, we obtain
\begin{equation}
\label{eq:fullloss}
\begin{aligned}
  & ~~~~~ \mathcal{L}(F,G_{x2y},G_{y2x},D_x,D_y) \\
  &= \mathcal{L}_{cdt}(G_{x2y},G_{y2x},D_x,D_y) + \lambda_{sm} \mathcal{L}_{sm}(F) \\
  &+ \lambda_{fx} \mathcal{L}_{fx}(G_{x2y},G_{y2x}) + \lambda_{fy} \mathcal{L}_{fy}(F) \\
  &+ \lambda_{corr} \mathcal{L}_{corr}(G_{x2y},G_{y2x}) + \lambda_{ms} \mathcal{L}_{ms}(G_{x2y}),
\end{aligned}
\end{equation}
where $\lambda_{s}, s\in \{sm, fx, fy, corr, ms\}$ weigh the relative importance among different objectives. We would discuss the effectiveness of each objective in Section~\ref{experiment} by ablation study. Our final goal is to solve the following optimization problem:
\begin{equation}
\label{eq:optim}
\begin{aligned}
  \max_{D_x,D_y} \min_{F,G_{x2y},G_{y2x}} \mathcal{L}(F,G_{x2y},G_{y2x},D_x,D_y).
\end{aligned}
\end{equation}

\section{Experiment}
\label{experiment}
\subsection{Implementation Detials}
\noindent \textbf{Network and training}. We adopt the architecture for our generator and dicriminator networks from CycleGAN~\cite{CycleGAN2017} with patch discriminator~\cite{pix2pix} and take DispNet~\cite{dispnet} as our stereo matching network. We implement this method on \textit{Pytorch}. For training our proposed joint domain translation and stereo matching framework, we partition the training into two stages. In the warm-up stage, we first train the domain translation networks with only $\mathcal{L}_{cdt}$ and $\mathcal{L}_{fx}$ for $10$ epochs, using Adam optimizer~\cite{adam} with the momentum $\beta_1=0.5, \beta_2=0.999$ and learning rate $\alpha=0.0002$. Then we train the stereo matching network with only $\mathcal{L}_{sm}$ for $50$ epochs, using Adam optimizer with the momentum $\beta_1=0.9, \beta_2=0.999$ and learning rate $\alpha=0.0001$. In the second stage, we train these two components together in an end-to-end manner and maintain the hyper-parameters unchanged. We alternatively optimize domain translation nets and stereo matching net with the full objective. We empirically set the trade-off factors as $\lambda_{cyc}=10$, $\lambda_{sm}=1$, $\lambda_{fx}=5$, $\lambda_{fy}=5$, $\lambda_{corr}=1$ and $\lambda_{ms}=0.1$.

\noindent \textbf{Datasets}. We take three datasets to testify the effectiveness of our proposed method. Two of them are synthetic datasets and the last one is real dataset. The first is Driving, a subset of a large synthetic dataset Sceneflow~\cite{dispnet}, which describes a virtual-world car driving scene. It contains fast sequences and slow sequences with both forward driving and backward driving scenes, the number of images summing up to $4,400$ totally. The image size in this dataset is $540 \times 960$ and the range of disparity value is $0-300$. The second is Synthia-SF~\cite{Ros_2016_CVPR}, which contains $6$ sequences featuring different scenarios and traffic conditions. There are $2,224$ images with associated ground-truth disparity maps. The image size is $1080\times 1920$ and its range of disparity is similar to Driving dataset. The last real dataset is KITTI2015~\cite{kittidataset}, containing 200 training images collected in real scenarios. Its image size is around $385\times 1242$ with disparity ranging from $0$ to around $180$. Due to the inconsistency of object size between Synthia-SF and KITTI2015, we resize all images in Synthia-SF to half and the corresponding disparity value is divided by $2$.

\noindent \textbf{Evaluation metrics}. We testify the effectiveness of our proposed method by the following evaluation metrics. End-point error (EPE) is the mean average disparity error in pixels. D1-all means the percentage of pixels whose absolute disparity error is larger than $3$ pixels or $5\%$ of ground-truth disparity value. Percentages of erroneous pixels larger than $2, 4, 5$ are reported. All these evaluation metrics are calculated for both non-occluded (Noc) and all (All) pixels. The inference time on single TITAN-X GPU is also recorded.

\begin{table*}
\begin{center}
\begin{tabular}{c|c|c|c|c|c|c|c|c|c|c|c}
\hline
\multirow{2}{*}{Dataset} & Ablation & \multicolumn{2}{c|}{D1-all (\%)} & \multicolumn{2}{c|}{EPE} & \multicolumn{2}{c|}{\textgreater 2px (\%)} & \multicolumn{2}{c|}{\textgreater 4px (\%)}  & \multicolumn{2}{c}{\textgreater 5px (\%)}    \\
\cline{3-12}
 & objective & Noc & All & Noc & All & Noc & All & Noc & All & Noc & All \\
\hline
& w/o $\mathcal{L}_{corr}$ & $8.95$ & $9.45$ & $1.532$ & $1.675$ & $16.00$ & $16.36$ & $5.88$ & $6.32$ & $4.65$ & $5.20$ \\
Synthia& w/o $\mathcal{L}_{fx}$ & $9.46$ & $10.02$ & $1.570$ & $1.706$ & $16.89$ & $17.13$ & $6.34$ & $6.79$ & $4.98$ & $5.43$ \\
to & w/o $\mathcal{L}_{fy}$ & $9.32$ & $9.89$ & $1.552$ & $1.690$ & $16.73$ & $16.95$ & $6.20$ & $6.62$ & $4.84$ & $5.31$\\
KITTI2015& w/o $\mathcal{L}_{ms}$ & $9.04$ & $9.53$ & $1.538$ & $1.668$ & $16.13$ & $16.48$ & $5.94$ & $6.43$ & $4.72$ & $5.26$\\
\cline{2-12}
& full obj. & $\textbf{8.78}$ & $\textbf{9.26}$ & $\textbf{1.488}$ & $\textbf{1.631}$ & $\textbf{15.74}$ & $\textbf{16.09}$ & $\textbf{5.73}$ & $\textbf{6.17}$ & $\textbf{4.55}$ & $\textbf{5.08}$ \\
\hline
\hline
& w/o $\mathcal{L}_{corr}$ & $25.64$ & $26.16$ & $2.633$ & $2.804$ & $39.88$ & $40.96$ & $16.46$ & $17.12$ & $11.57$ & $12.11$ \\
Driving& w/o $\mathcal{L}_{fx}$ & $26.38$ & $26.95$ & $2.883$ & $3.029$ & $40.61$ & $41.64$ & $17.28$ & $17.96$ & $12.64$ & $13.28$ \\
to & w/o $\mathcal{L}_{fy}$ & $26.22$ & $26.79$ & $2.843$ & $2.998$ & $40.42$ & $41.28$ & $17.06$ & $17.85$ & $12.09$ & $12.66$ \\
KITTI2015& w/o $\mathcal{L}_{ms}$ & $25.45$ & $25.98$ & $2.601$ & $2.782$ & $39.76$ & $40.85$ & $16.30$ & $16.95$ & $11.34$ & $11.93$ \\
\cline{2-12}
& full obj. & $\textbf{25.18}$ & $\textbf{25.71}$ & $\textbf{2.584}$ & $\textbf{2.752}$ & $\textbf{39.16}$ & $\textbf{40.24}$ & $\textbf{15.83}$ & $\textbf{16.55}$ & $\textbf{11.04}$ & $\textbf{11.60}$ \\
\hline
\end{tabular}
\end{center}
\caption{Evaluation results of the proposed method with different objectives by ablation study. Lower value means better performance. }
\label{table:ablation}
\end{table*}

\subsection{Comparison with Other Methods}
We first investigate whether the proposed method is superior to other related methods or not, whose results are summarized in Table~\ref{table:comparison}. We take two synthetic data - Synthia and Driving as our source-domain dataset, and one real dataset - KITTI2015 as our target-domain dataset. A dubbed method without domain translation, which is called Inference, is to train the stereo matching network on synthetic data and then directly predict disparity map on real data. Two state-of-the-art unsupervised adaptation methods for stereo matching are compared. Particularly, we use SL+Ad to denote unsupervised online adaptation method described in~\cite{Tonioni_2019_realadapt} and use L2A+Wad to denote unsupervised adaptation via meta learning framework described in~\cite{Tonioni_2019_learn2adapt}. Moreover, since there is no stereo matching-specific domain adaptation technique developed, we choose CycleGAN~\cite{CycleGAN2017} as our baseline for comparison. For the sake of fair comparison, we set the stereo matching network of all methods to DispNet~\cite{dispnet}.

As could be seen from Table~\ref{table:comparison}, all of the methods perform better on Synthia-to-KITTI2015 than Driving-to-KITTI2015 because there is a larger gap between Driving and KITTI2015. Among these methods, Inference perform worst due to the natural gap between synthetic and real domain. SL+Ad updates the stereo matching network by calculating the error between the inversely-warped left image and real left image when accessing new stereo images. L2A+Wad proposes a novel weight confidence-guided adaptation technique and updates the network in a meta-learning manner. These two methods mitigated the domain gap to a little bit extent but meanwhile brought some extra calculation burden to inference process. Their inference time increase from $0.06$ seconds to $0.19$ and $0.23$ seconds respectively. The translation results of CycleGAN have the problem of pixel distortion, as introduced in Section~\ref{introduction}, so it performed not well enough. The proposed joint domain translation and stereo matching framework, with novel stereo constraints, beat all the above methods by reducing the number of erroneous pixels considerably. The significant improvements in all evaluation metrics demonstrate the superiority of our method. In addition, the inference time of our method is same as that of original DispNet because all the extra domain translation and auxiliary training is completed in the procedure of offline training.

\subsection{Ablation Study}

We then investigate how each objective term influence the performance of unsupervised stereo matching quantitatively by ablation study. Besides cycle domain translation loss and stereo matching loss, we propose four novel objectives for regularizing the basic problem formulation including correlation consistency loss, mode seeking loss, feature re-projection loss for real stereo and for synthetic stereo. We would train our joint framework by removing one of them and then record the corresponding D1-all, EPE, and bad pixel percentage with threshold $2$, $4$ and $5$, as summarized in Table~\ref{table:ablation}. The results of ablation study on both Synthia and Driving source dataset show similar trend. In general, feature re-projection loss for synthetic stereo and real stereo is more effective than that of correlation consistency loss and mode seeking loss. We try to analyze the reasons in the following.

First of all, among all four proposed objectives, feature re-projection loss for synthetic stereo $\mathcal{L}_{fx}$ is most effective on our joint framework. The reasons are as follows: 1) it ensures that translated outputs be stereo-consistent with inputs, which is vital to stereo matching loss in the presence of a large amount accurate disparities; 2) it benefits the training of stereo matching network with feature re-projection loss for real stereo by well-learned translation networks.

The effect of feature re-projection loss for real stereo $\mathcal{L}_{fy}$ is runner-up, because it actually provides extra training signals for training stereo matching network. However, such supervision signals are obtained from the warping of features in domain translation networks, so its performance is highly dependent on how well domain translation networks are trained by $\mathcal{L}_{fx}$ to a large degree.

Thirdly, correlation consistency loss $\mathcal{L}_{corr}$ may contribute to this framework marginally in the presence of feature re-projection losses. It serves as a complement to $\mathcal{L}_{fx}$. As analyzed above, feature re-projection loss for real stereo images usually benefits from the well-trained translation networks by $\mathcal{L}_{fx}$. However, sometimes the value of feature re-projection loss for real stereo images may be low, but contrarily, both pixel distortion in translation and inaccurate estimation in stereo matching occur simultaneously. This correlation consistency loss could help only at this time.

Finally, D1-all results would drop a little bit without mode seeking loss. Because mode seeking loss actually provides fine-grained diversity to translated results and essentially helps stereo matching network learn a more robust disparity estimation network. In other words, stereo matching networks would learn to reduce the influence of various noise and lighting conditions during training.

Thanks to the integration of all the above four objectives described in Equation~\ref{eq:fullloss}, we have obtained great improvement on filling the synthetic-to-real gap in stereo matching.

\subsection{The Effect of Stereo Matching Network}
\begin{table}
\begin{center}
\begin{tabular}{l|c|c|c|c|c}
\hline
    \multicolumn{6}{c}{Synthia-to-KITTI2015} \\
\hline
\multirow{2}{*}{Models}  & \multicolumn{2}{c|}{Inference} & \multicolumn{2}{c|}{Proposed} & Time  \\
\cline{2-5}
    & D1-all & EPE & D1-all & EPE & (s) \\
\hline
DispNet & $11.14$ & $1.961$ & $9.26$ & $1.631$ & $0.06$ \\
\hline
GwcNet~\cite{guo2019group} & $7.46$ & $1.576$ & $5.74$ & $1.424$ & $0.32$ \\
\hline
\hline
    \multicolumn{6}{c}{Driving-to-KITTI2015} \\
\hline
\multirow{2}{*}{Models}  & \multicolumn{2}{c|}{Inference} & \multicolumn{2}{c|}{Proposed} & Time  \\
\cline{2-5}
    & D1-all & EPE & D1-all & EPE & (s) \\
\hline
DispNet & $53.07$ & $9.513$ & $25.71$ & $2.752$ & $0.06$ \\
\hline
GwcNet~\cite{guo2019group} & $28.21$ & $3.275$ & $12.17$ & $1.980$ & $0.32$ \\
\hline
\end{tabular}
\end{center}
\caption{The effect of different stereo matching network. Lower value means better performance. }
\label{table:smnet}
\end{table}

In this part, we show how the structure of stereo matching network influences the performance of our proposed joint domain translation and stereo matching framework. We compare DispNet with one of the recently-proposed state-of-the-art stereo matching model GwcNet~\cite{guo2019group}. Their D1-all and EPE scores and inference time are reported in Table~\ref{table:smnet}. As can be seen, GwcNet~\cite{guo2019group} performs far better than DispNet on both datasets and evaluation metrics. When using Synthia as our synthetic training data, our proposed model could help DispNet reduce D1-all and EPE by around $16.8\%$. It also makes GwcNet reduce D1-all by $23\%$ and reduce EPE by $9.6\%$. For Driving training data whose domain gap to KITTI2015 is larger, our method could also help stereo matching network obtain very competitive performance. After trained with our proposed framework, D1-all is reduced by $51.5\%$ and EPE $71\%$ for DispNet respectively and D1-all is reduced by $56.8\%$ and EPE by $39.6\%$ for GwcNet respectively.

\subsection{Generalization to Other Real Datasets}
\begin{table}
\begin{center}
\begin{tabular}{l|c|c|c|c}
\hline
    \multicolumn{5}{c}{D1-all} \\
\hline
    \multirow{2}{*}{\backslashbox{Train}{\kern-2em Test}} & \multicolumn{2}{c|}{KITTI2012} & \multicolumn{2}{c}{Cityscapes} \\
\cline{2-5}
    & Inference & Proposed & Inference & Proposed \\
\hline
Synthia & $13.34$ & $11.56$ & $31.69$ & $22.93$ \\
\hline
Driving & $56.31$ & $25.57$ & $60.50$ & $32.14$ \\
\hline
\hline
    \multicolumn{5}{c}{EPE} \\
\hline
    \multirow{2}{*}{\backslashbox{Train}{\kern-2em Test}} & \multicolumn{2}{c|}{KITTI2012} & \multicolumn{2}{c}{Cityscapes} \\
\cline{2-5}
    & Inference & Proposed & Inference & Proposed \\
\hline
Synthia & $2.121$ & $1.936$ & $11.805$ & $6.701$ \\
\hline
Driving & $11.669$ & $2.832$ & $15.468$ & $8.506$ \\
\hline
\end{tabular}
\end{center}
\caption{Generalization capability of our proposed method. We test our performance on two other real dataset: KITTI2012 and Cityscapes. Models are trained with only synthetic dataset and KITTI2015 dataset. }
\label{table:generalization}
\end{table}

To demonstrate the generalization capability of stereo matching network trained in our joint optimization framework, we test their performance on other two real datasets - KITTI2012~\cite{geiger2012we} and Cityscapes~\cite{Cordts2016Cityscapes}, whose results are summarized in Table~\ref{table:generalization}. Images in KITTI2012 have very similar domain style to those in KITTI2015 due to their similar camera setting. Therefore, the performance gain with the help of domain translation on KITTI2012 is similar to that on KITTI2015. For Cityscapes real dataset, both D1-all and EPE scores almost reduce by half. These significant improvements demonstrate great generalization capability of our proposed joint framework.

\section{Conclusion and Future Work}
In this paper we propose a novel end-to-end framework that trains domain translation networks and stereo matching network jointly. The newly-introduced stereo constraints including correlation consistency loss, bi-directional multi-scale feature re-projection loss and mode seeking loss regularize this joint framework to achieve better performance on stereo matching without ground-truth. The experimental results testify the effectiveness of our proposed framework in bridging the synthetic-to-real domain gap.

Our proposed framework successfully mitigated the gap between synthetic and real domain, yet there usually exist other gaps on intrinsics and disparity distribution between real-domain stereo images and translated-real stereo images, which is not explicit in our experimental datasets. Further study is also required to facilitate the generalization capability of our framework when meeting such datasets.

\noindent \footnotesize{\textbf{Acknowledgement}. This work is supported in part by SenseTime Group Limited, and in part by the General Research Fund through the Research Grants Council of Hong Kong under Grants CUHK14202217, CUHK14203118, CUHK14205615, CUHK14207814, CUHK14213616, CUHK14207319, CUHK14208619, and in part by Research Impact Fund R5001-18.}

{\small
\bibliographystyle{ieee_fullname}

}

\end{document}